\documentclass[lettersize,journal]{IEEEtran}
\usepackage{geometry}
\usepackage[font=footnotesize]{caption}

\usepackage{amsmath,amsfonts}
\usepackage{algorithmic}
\usepackage{algorithm}
\usepackage{array}
\usepackage{textcomp}
\usepackage{stfloats}
\usepackage{url}
\usepackage{verbatim}
\usepackage{graphicx}
\usepackage{xcolor}
\usepackage{cite}
\usepackage{gensymb}

\definecolor{mygray}{gray}{0.9}

\usepackage{verbatim}

\usepackage[table]{xcolor}

\usepackage{balance}

\usepackage{booktabs}

\usepackage{subfig}
\usepackage{subcaption}

\begin{document}

\title{\fontsize{22}{26}\selectfont LiDAR for Crowd Management: Applications, Benefits, and Future Directions}

\author{Abdullah Khanfor, Chaima Zaghouani, Hakim Ghazzai, Ahmad Alsharoa, and Gianluca Setti

\thanks{\newline
\rule{\linewidth}{0.05pt}\\
Abdullah Khanfor is with the College of Computer Science and Information Systems, Najran University, Najran, Saudi Arabia.\\
Chaima Zaghouani and Ahmad Alsharoa are with the College of Innovation \& Technology, University of Michigan-Flint, MI, USA.\\
Hakim Ghazzai and Gianluca Setti are with King Abdullah University of Science and Technology (KAUST), Thuwal, Saudi Arabia.\\
\textcolor{mygray}{\rule{\linewidth}{0.4pt}}\\
This paper is accepted for publication in the IEEE Internet of Things Magazine, Mar. 2026. \newline \textcopyright~2026 IEEE. Personal use of this material is permitted. Permission from IEEE must be obtained for all other uses, in any current or future media, including reprinting/republishing this material for advertising or promotional purposes, creating new collective works, for resale or redistribution to servers or lists, or reuse of any copyrighted component of this work in other works.}
}
\maketitle
\thispagestyle{empty}
\pagestyle{empty}

\begin{abstract}
Light Detection and Ranging (LiDAR) technology offers significant advantages for effective crowd management. This article presents LiDAR technology and highlights its primary advantages over other monitoring technologies, including enhanced privacy, performance in various weather conditions, and precise 3D mapping. We present a general taxonomy of four key tasks in crowd management: crowd detection, counting, tracking, and behavior classification, with illustrative examples of LiDAR applications for each task. We identify challenges and open research directions, including the scarcity of dedicated datasets, sensor fusion requirements, artificial intelligence integration, and processing needs for LiDAR point clouds. This article offers actionable insights for developing crowd management solutions tailored to public safety applications.
\end{abstract}


\section{Introduction}

\IEEEPARstart{E}{fficient} crowd management is essential for many smart city applications, supporting public safety, transportation planning, and event control. Timely insights and actions are crucial for monitoring pedestrian flow and managing crowds in public spaces~\cite{hamrouni2023multi,10942376}. Without precise management, crowd events can result in injuries and fatalities. 
One of the most recent crowd stampedes occurred in February 2025 in a New Delhi railway station, with at least 18 casualties. These incidents are not limited to specific events or areas and often result from inadequate crowd management practices. 

Traditional methods, such as police presence, physical barriers, and manual supervision, remain widely used, but they are labor-intensive and often reactive. Recent technological advances can help mitigate these risks and support informed, life-saving decisions. IoT sensors, Wi-Fi signals, and camera-based systems are investigated to enable more proactive crowd monitoring; however, these approaches often raise concerns around privacy, scalability, and efficiency, particularly when aiming for high precision in fully automated systems.

Light detection and ranging (LiDAR) sensors can serve as an alternative and complementary technology that provides accurate, privacy-preserving spatial sensing. Unlike cameras, LiDAR sensors perceive the environment by emitting pulsed light beams. These beams reflect off surrounding objects and return to the sensor, allowing the system to compute precise distances. The resulting reflections are stored as a 3D point cloud, with each point representing a location in space. This enables the creation of a detailed and highly accurate 3D model of the environment, making LiDAR especially valuable for real-time crowd monitoring and detection without capturing identifiable personal data.


LiDAR technology has been widely adopted in autonomous ground vehicles for object detection and distance estimation applications. The use of LiDAR in vehicular systems began in the early 2000s and gained visibility through the 2005 DARPA Grand Challenge, where it demonstrated effectiveness in autonomous navigation tasks. 
Several major automotive manufacturers have integrated LiDAR systems for autonomous driving and driver assistance features. Recently, researchers have explored LiDAR applications beyond vehicle navigation, particularly in monitoring human activities. The technology's precision enables the detection of subtle physiological changes, including respiratory patterns reflected in chest wall movements, allowing non-contact measurement of vital signs. This capability suggests potential applications in healthcare monitoring and human activity recognition systems~\cite{rinchi2023lidar,10472936}. Applications include human activity recognition in indoor contexts~\cite{meng2025indoor} and medical monitoring, achieving over 93\% accuracy for healthcare activities~\cite{10472936}.



\begin{figure}[!t]
\centering
\includegraphics[width=\columnwidth]{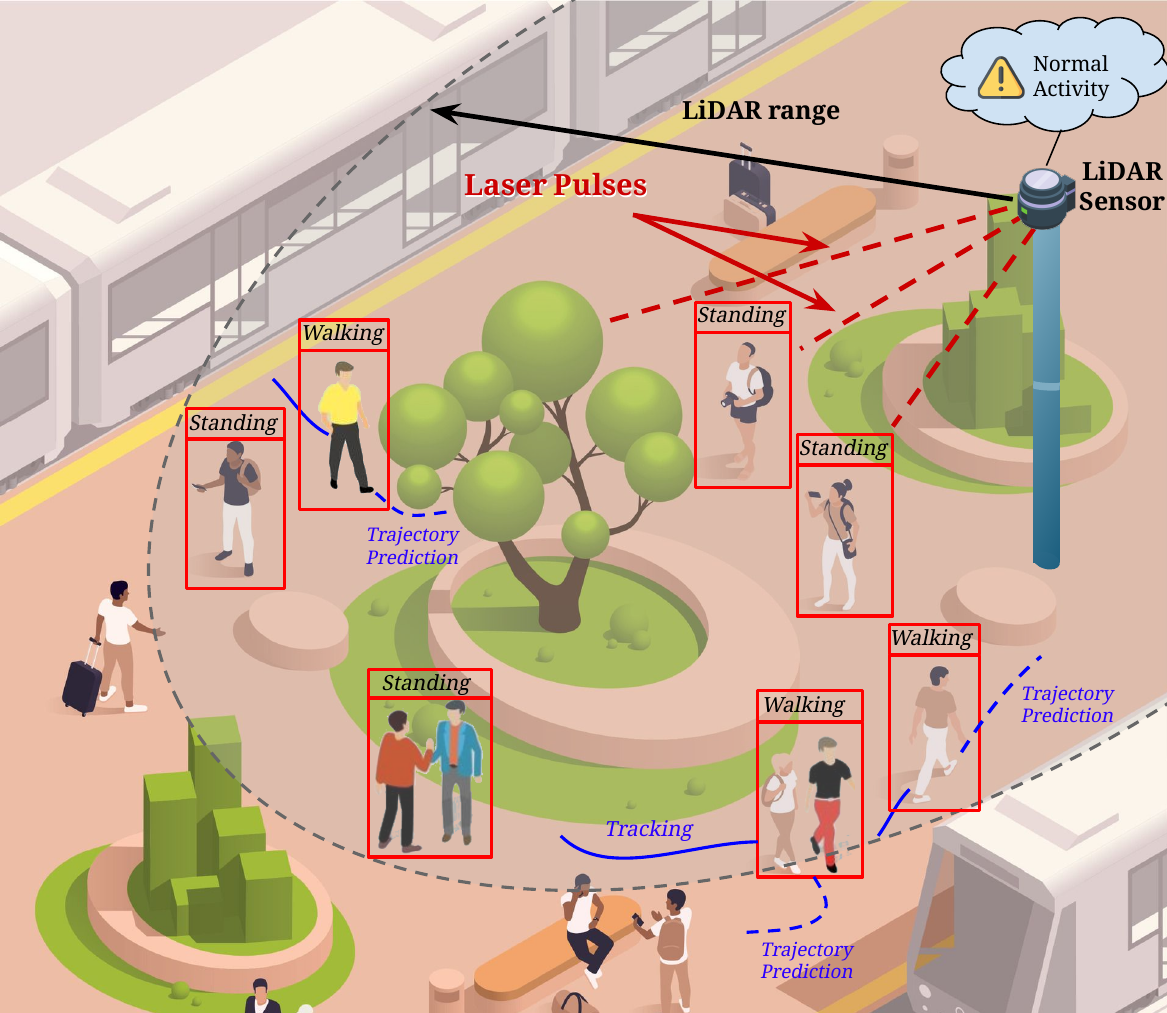}
\caption{Illustration of LiDAR scanning the crowd in public space.}
\label{fig_uavLidar}
\end{figure}



This article investigates LiDAR's potential for crowd management applications. We identify LiDAR's benefits, discuss LiDAR-based crowd descriptors, highlight applications of LiDAR technology for crowd detection, counting, tracking, and behavior classification, and provide a taxonomy of implementation challenges and open research directions.

    

\begin{figure*}[!t]
    \centering
    \begin{minipage}{0.48\textwidth}
        \centering
        \includegraphics[width=\textwidth]{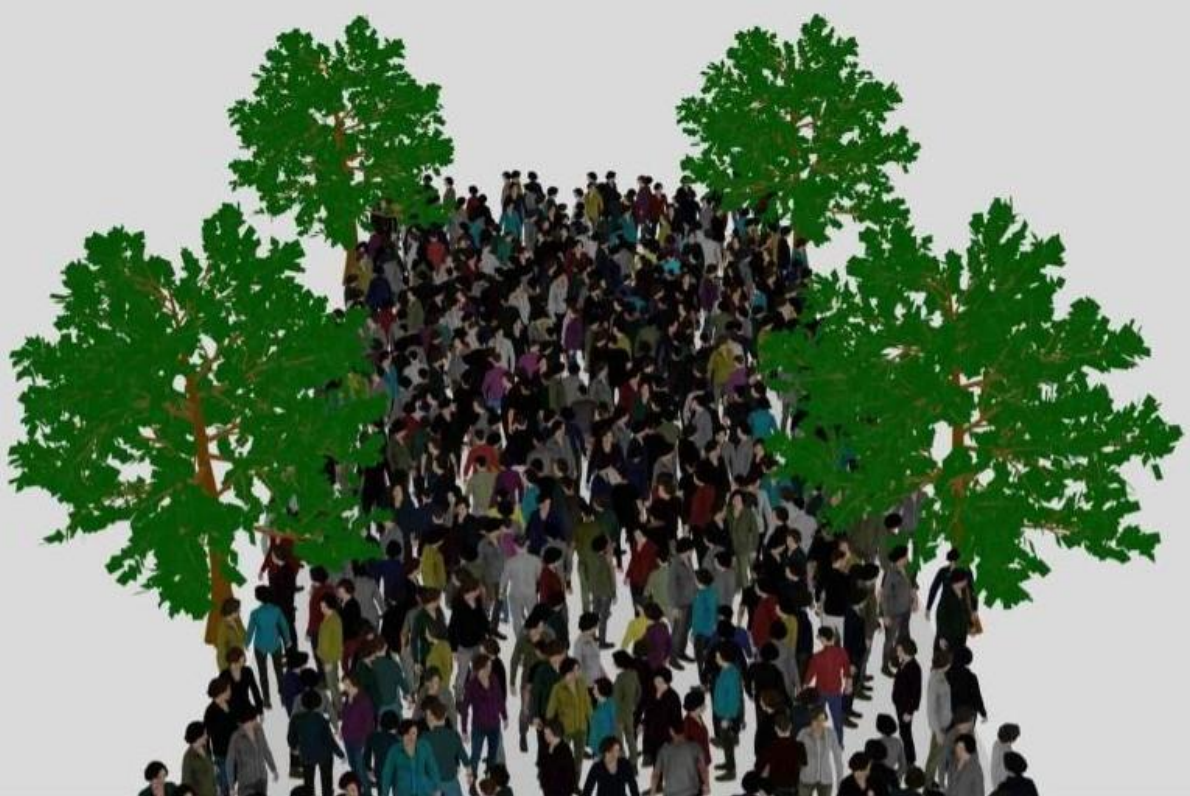}\\[0.2cm]
        \textbf{(a) RGB Image}
    \end{minipage}
    \hfill
    \begin{minipage}{0.48\textwidth}
        \centering
        \includegraphics[width=\textwidth]{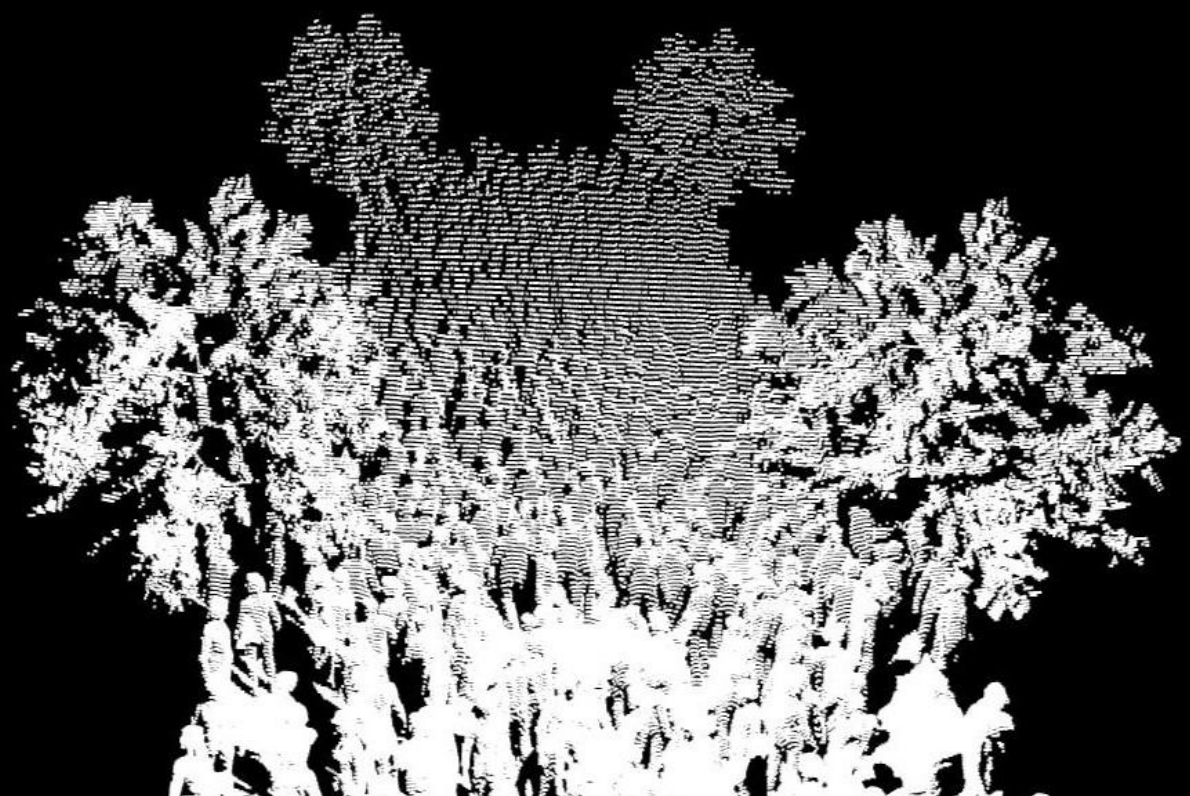}\\[0.2cm]
        \textbf{(b) Point Cloud}
    \end{minipage}
    
    \caption{Comparison of an RGB camera image versus a LiDAR point cloud for the same crowded scene generated using a simulator. LiDAR eliminates people's faces and identifying features from collected data, preserving individual privacy in monitored areas.}
    \label{fig:rgb_vs_lidar}
\end{figure*}

\section{Benefits of LiDAR for Crowd Management}
\textcolor{black}{LiDAR technology offers distinct advantages that make it well-suited for crowd management. It provides rich spatial data with high precision, operates reliably under varying environmental conditions, and inherently preserves individual privacy. These characteristics enable LiDAR to serve as a powerful sensing modality for monitoring and analyzing crowds, particularly in complex and dynamic public environments. In the following, we highlight the primary benefits of LiDAR technologies for crowd management, while a comparative discussion with other sensing modalities and their respective trade-offs is provided separately in Table~\ref{tab_compare}.}

\subsection{3D Point Cloud, Precision, and Reflectivity}
\textcolor{black}{LiDAR’s most significant strength lies in its ability to generate precise 3D point clouds that capture the geometric structure of the environment, enabling accurate localization of pedestrians and crowd formations, as well as reliable spatial awareness and object separation. In addition to spatial coordinates, LiDAR captures reflectivity values that help distinguish materials such as clothing, skin, and infrastructure surfaces, thereby improving segmentation and tracking performance. The resulting 3D spatial understanding enables effective crowd flow monitoring, bottleneck detection, and behavior analysis in complex environments.} 

\subsection{Extended Coverage and Variable Fields of View}
\textcolor{black}{Modern LiDAR systems provide extended detection ranges, often exceeding 200 meters, along with wide or full 360-degree fields of view. 
While Wide Field of View (FOV) cameras and multi-camera systems can achieve comparable coverage, LiDAR provides direct 3D measurements that simplify occlusion handling without requiring complex multi-sensor calibration. For example, the Ouster OS2-128 offers a detection range of up to 240 meters with a 360-degree horizontal and 22.5-degree vertical FOV, enabling it to monitor several thousand square meters with fine-grained resolution. In practical terms, this allows tracking crowd movement and density across an entire public square, stadium perimeter, or transport terminal with a single or a few mounted units. Such coverage not only reduces installation and maintenance costs but also enhances situational awareness across large public spaces.}

\subsection{Robust Performance Under Varying Conditions}
\textcolor{black}{LiDAR performance is largely independent of ambient lighting conditions, enabling reliable operation in bright daylight, under direct sunlight, and in complete darkness. Because it does not rely on passive image formation, LiDAR avoids issues such as lens flare or image washout that can affect optical systems. This characteristic supports consistent data quality across a wide range of environments and operating times, making LiDAR well-suited for continuous crowd monitoring. Moreover, LiDAR’s reflection-based sensing maintains stable performance in light fog, rain, and snow conditions under which purely vision-based approaches may experience degradation.}

\subsection{Privacy Preservation}
\textcolor{black}{LiDAR inherently supports privacy-aware monitoring, as it does not capture personally identifiable information. The resulting point cloud data consists of geometric representations without facial features, skin tone, or fine-grained textual content, such as information displayed on screens or documents, thereby safeguarding individual anonymity, as illustrated in Fig.~\ref{fig:rgb_vs_lidar}. While camera systems can implement privacy measures at the processing level, LiDAR provides inherent privacy preservation at the sensing level.}

\subsection{Data Manipulation and Integration}
\textcolor{black}{LiDAR-generated point clouds are highly structured, making them well-suited for downstream processing and integration with AI pipelines. The data can be efficiently cropped, clustered, voxelized, classified, and fused with other modalities such as radar or camera data. This flexibility supports a wide range of crowd management tasks, including pedestrian detection, activity recognition, and abnormal behavior analysis, and facilitates the integration of LiDAR data into end-to-end intelligent crowd monitoring systems.}\\

\textcolor{black}{While this section focuses on the key benefits of LiDAR for crowd management, Table~\ref{tab_compare} provides a comparative overview of LiDAR, RGB, and thermal cameras across key attributes relevant to crowd management. The comparison highlights the different sensing properties and operational trade-offs of each modality, including LiDAR’s native 3D spatial perception and privacy-aware sensing, along with the associated computational and deployment considerations. Other challenges related to LiDAR usage for crowd management are further discussed in Section~\ref{sec:directions}.}

\begin{table*}[!t]
\rowcolors{2}{gray!10}{white}
\centering
\caption{\textcolor{black}{Technical and Operational Comparison of Crowd Management Sensors}}
\label{tab_compare}
\begin{tabular}{ p{2.6cm} p{4.5cm} p{4.5cm} p{4.5cm} }
\toprule
\textbf{Feature} & \textbf{LiDAR} & \textbf{RGB Camera} & \textbf{Thermal Camera} \\
\midrule
\textbf{3D capability} 
& Native 3D point cloud 
& 2D images; depth via stereo or learning-based methods 
& 2D thermal map; limited depth perception \\

\addlinespace
\textbf{Field of view} 
& Wide (up to 360$^\circ$ depending on sensor model) 
& Wide coverage achievable using fisheye lenses or multi-camera setups 
& Limited by lens (typically 40–90$^\circ$) \\

\addlinespace
\textbf{Detection range} 
& Up to 200–300 m (high-resolution LiDAR) 
& 50–150 m (lens, resolution, and lighting dependent) 
& 10–50 m (sensor dependent) \\

\addlinespace
\textbf{Distance accuracy} 
& High – direct range measurement (cm-level) 
& Moderate – depth inferred indirectly 
& Low to moderate – coarse distance estimation \\

\addlinespace
\textbf{Illumination requirement} 
& None – active sensing, operates in darkness 
& Requires sufficient lighting or IR assistance 
& None – operates in darkness \\

\addlinespace
\textbf{Privacy} 
& High – inherent privacy at sensing level 
& Privacy achievable via system-level techniques (e.g., blurring, on-device analytics) 
& Medium – body silhouettes without facial detail \\

\addlinespace
\textbf{Weather resistance} 
& Moderate to high – degraded in heavy rain or dense fog 
& Low – sensitive to weather and illumination changes 
& Moderate – robust in darkness, affected by ambient heat \\

\addlinespace
\textbf{Computational requirements} 
& High – point cloud processing and 3D analytics 
& Medium – image processing; additional load for depth or privacy 
& Low – simple thermal analysis \\

\addlinespace
\textbf{Deployment cost} 
& High – sensor cost and calibration effort 
& Low to medium – inexpensive sensors; higher cost for multi-camera systems 
& Medium – moderate sensor and integration cost \\

\addlinespace
\textbf{Data storage \& processing} 
& High – large point clouds and specialized pipelines 
& Medium – image data; increased for multi-view setups 
& Low – compact data and simple processing \\
\bottomrule
\end{tabular}
\end{table*}

\section{LiDAR-based Crowd Descriptors}


LiDAR sensors generate 3D point clouds that capture both the geometric structure and temporal dynamics of crowds. \textcolor{black}{Unlike structured image data, the raw output of a LiDAR sensor consists of unordered sets of 3D coordinates $(x, y, z)$ and an associated intensity (or reflectance) value corresponding to the returned laser signal. This irregular, sparse, and high-dimensional data structure lacks the spatial grid organization of images, requiring specialized processing algorithms that handle variable point densities and maintain spatial relationships without relying on pixel-based representations. Together, the geometric and intensity information enables the extraction of multidimensional descriptors that are essential for detection, tracking, and behavior analysis in crowd management scenarios.}

\subsection{Geometric Descriptors}
Geometric descriptors extracted from LiDAR point clouds provide critical features for pedestrian discrimination in dense crowds. Human subjects typically exhibit heights between 1.5 and 2.0 meters, with recognizable geometric signatures corresponding to anatomical structures such as head, body, hands, and legs. The aspect ratios of fitted bounding boxes help distinguish individuals within crowded areas of interest. Surface normals computed from local point neighborhoods reveal object boundaries and orientations, while principal curvatures, computed via eigenvalue analysis, differentiate convex human forms from planar surfaces. These geometric descriptors remain invariant to lighting conditions, providing consistent detection capabilities during both day and night operations. \textcolor{black}{As an example, PointNet++ 
exploits such geometric descriptors through hierarchical feature learning, aggregating local surface and shape information across multiple spatial scales to form robust representations for pedestrian classification.}

\subsection{Spatial Distribution Descriptors}
Statistical analysis of point cloud distributions reveals characteristic spatial patterns in crowd formations, supporting real-time monitoring and risk assessment. Interpersonal spacing, computed as the minimum distance between neighboring individuals, is a fundamental indicator of local crowd density, with small distances signaling congestion and potential safety risks. To efficiently capture these spatial relationships, the monitored area can be partitioned using Voronoi tessellation, where cell areas inversely correlate with local density while avoiding exhaustive pairwise distance computations. In addition, spatial statistics such as Ripley’s $K$ function can be extended to 3D point clouds to quantify clustering behavior and distinguish between structured formations (e.g., queues) and random crowd distributions. \textcolor{black}{For instance, graph neural networks can directly exploit these spatial descriptors by representing individuals as nodes and interpersonal distances or Voronoi adjacency as edges, enabling learning-based modeling of crowd density and spatial interaction dynamics.}

\subsection{Temporal Descriptors}
Temporal analysis of LiDAR point cloud sequences provides key insights into crowd dynamics by capturing position changes across consecutive frames. Motion descriptors such as velocity and direction can be estimated using optical flow techniques adapted to 3D data, while iterative closest point (ICP) algorithms are commonly employed to register successive point clouds and minimize alignment errors. These approaches enable the detection of transitions from smooth, laminar crowd motion to irregular or turbulent patterns that may indicate emerging safety risks. By analyzing temporal coherence and motion evolution, anomalous behaviors and sudden changes in crowd dynamics can be identified in real time. \textcolor{black}{In this context, recurrent neural networks, such as LSTM architectures, can process sequences of temporal descriptors to learn typical crowd flow patterns and predict trajectory deviations or abnormal motion trends.}

\subsection{Reflectance-based Descriptors}
LiDAR-intensity values provide complementary information for crowd analysis beyond geometric structure. Variations in returned intensity can reveal differences in surface properties and clothing materials, supporting improved separation between individuals and surrounding infrastructure. Modern LiDAR sensors with multi-echo capabilities can further capture object contours and semi-transparent surfaces~\cite{man2021multi}, while overlapping laser returns from densely packed individuals can serve as indicators of crowd compactness. These reflectance-based descriptors enrich point cloud representations and enhance robustness in complex scenes. \textcolor{black}{For example, PointCNN~\cite{NEURIPS2018_f5f8590c} can effectively incorporate intensity values via convolution-like operations on point sets, thereby improving person–object discrimination in challenging crowd environments.}

\subsection{Statistical and Structural Descriptors}
Statistical analysis of point cloud distributions reveals informative patterns for crowd management applications. Higher-order statistical moments, such as skewness and variance, can reflect variations in individual heights and spatial dispersion, providing insights into demographic composition and potential visibility constraints. Beyond point-wise statistics, structural descriptors derived from spatial distributions capture persistent formation patterns that remain stable despite continuous individual movement within the crowd. These descriptors can be visualized through representations such as crowd density maps or structural heatmaps, enabling operators to monitor how formations emerge, stabilize, or dissolve over time. \textcolor{black}{For example, graph-based learning models and autoencoder architectures can leverage these statistical and structural descriptors to learn compact representations of crowd organization and detect deviations from typical formation patterns.}\\


These LiDAR-based descriptors provide the quantitative foundation for implementing the crowd management tasks discussed in the following section. Their 3D nature and independence from lighting conditions offer unique advantages, though converting this rich information into actionable insights remains an active area of research. \textcolor{black}{Jointly or individually, these descriptors are commonly used as input features for point cloud–based deep learning architectures. The coupling between descriptor design and learning architecture allows LiDAR-based AI models to effectively exploit both local geometric detail and global crowd dynamics.}

\section{LiDAR for Crowd Management Tasks}
\label{tasks}

\begin{figure*}[!t]
\centering
\includegraphics[width=\textwidth]{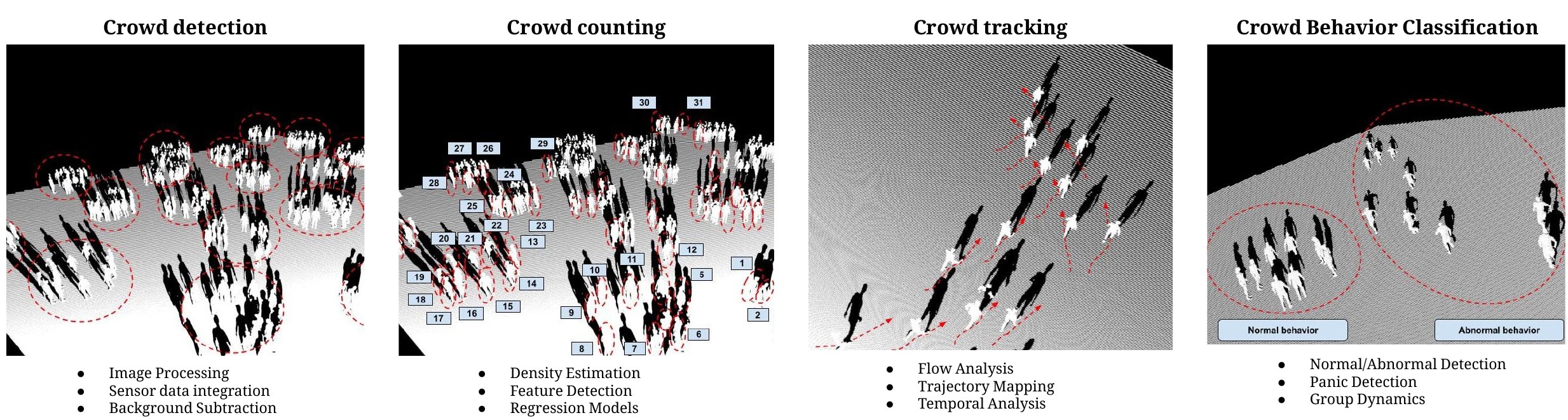}
\caption{Representation of crowd management tasks and the analytical tools necessary for implementing effective crowd control strategies.}
\label{fig_aspects}
\end{figure*}


Crowd management applications can be categorized into four primary tasks: detection, counting, tracking, and behavior analysis. Traditionally handled through manual observation, signage, and physical barriers, these tasks have increasingly benefited from advancements in sensing and computation. This section explores how LiDAR enables each of these tasks by leveraging its rich spatial and temporal descriptors to deliver reliable, scalable, and intelligent crowd management solutions.

\subsection{Crowd Detection}
\label{detection}
Crowd detection is the foundational task in crowd management, aiming to identify the presence and spatial distribution of individuals, groups, and entire crowd formations within a monitored area. Depending on the application, detection may focus on locating each individual (e.g., for counting or tracking), identifying specific subgroups such as children, women, or security personnel based on physical characteristics, or analyzing the structure and density of the crowd as a whole. Accurate detection at all three levels, individual, subgroup, and global, is critical for enabling higher-level tasks such as tracking, behavior analysis, and risk assessment.

LiDAR offers reliable capabilities for crowd detection by generating dense 3D point clouds that encode geometric and reflectivity information about the environment. Individual pedestrians can be identified as discrete point clusters based on features such as height, body shape, and aspect ratio. Groups within the crowd can be detected by clustering closely spaced individuals, with further analysis of spatial distributions or reflectance values to infer demographic characteristics or roles. For example, the relatively lower height and body mass of children may be inferred through bounding box analysis, while material differences in clothing may help distinguish subgroups when combined with reflectivity data. At the crowd level, LiDAR provides a holistic volumetric view, enabling detection of compact formations, bottlenecks, or high-density zones, even in environments with poor lighting or heavy occlusion.




\subsection{Crowd Counting}
\label{counting}
Crowd counting aims to estimate the number of people present in a given space, either as a total headcount or a localized density map. This task is essential for crowd flow regulation, event management, safety planning, and public space optimization. Counting can be performed at different granularities, globally across the entire scene, or locally within specific zones such as entrances, platforms, or exits. In highly dynamic or densely packed environments, traditional image-based counting approaches often struggle with occlusion, perspective distortion, and overlapping individuals, resulting in unreliable estimates.

By capturing high-resolution 3D point clouds, LiDAR sensors preserve the geometric structure of the scene. 
Crowd density can be computed using voxel-based occupancy grids, spatial clustering algorithms, or Voronoi partitioning, each offering real-time adaptability to changing crowd conditions. These techniques enable accurate counting in both sparse and highly congested areas while preserving privacy due to the absence of facial or identifying information.



Traditional image-based crowd counting has made significant advances with deep learning; however, these models require adaptation for effective application to 3D LiDAR data to successfully identify individuals and groups, enabling both counting and directional flow estimation from raw point cloud data. In this context, deep learning architectures for crowd tasks on point cloud data remain a challenging research problem, especially in highly dynamic and cluttered environments with varying sensing conditions and unconstrained settings~\cite{li2020deep}.

\subsection{Crowd Tracking}
\label{tracking}
Crowd tracking focuses on maintaining the spatiotemporal continuity of individuals or groups as they move through a monitored space. It enables real-time analysis of crowd flow, detection of directional trends, and identification of mobility bottlenecks. Effective tracking is essential for applications such as people counting over time, evacuation planning, and congestion mitigation. In dense environments, tracking becomes challenging due to frequent occlusions, intersecting trajectories, and dynamic scene changes.

LiDAR enhances tracking capabilities by providing consistent, frame-by-frame 3D representations of the environment. Individuals can be tracked by associating their corresponding point cloud clusters across consecutive frames using data association techniques such as Kalman filtering, probabilistic data association, or nearest-neighbor matching. More advanced methods leverage iterative closest point (ICP) algorithms or 3D optical flow to register motion fields and accurately estimate object trajectories. \textcolor{black}{In top-down or low-occlusion layouts, direct 3D spatial measurements can help reduce ID switches by limiting perspective-related ambiguities; however, in dense or highly self-occluding crowds, tracking performance may degrade, and camera-based appearance cues can provide complementary benefits, particularly at extended ranges where LiDAR spatial resolution decreases.}


\textcolor{black}{In addition to individual tracking, LiDAR supports crowd flow estimation by analyzing collective movement patterns and temporal variations in spatial density. By aggregating trajectories or directly modeling spatiotemporal point cloud dynamics, flow-level descriptors such as dominant movement directions, velocity fields, and congestion patterns can be extracted. Trajectory clustering and flow field modeling techniques enable the characterization of macroscopic crowd behaviors, including lane formation, wandering, merging, walking and standing gatherings, dispersal, splitting, and directional shifts~\cite{zhou2024crowd}. 
Recent studies have adapted multi-object tracking frameworks and 3D optical flow techniques to LiDAR point cloud data, enabling flow estimation and collective behavior analysis in crowd scenarios where direct visual appearance cues are limited.}



\subsection{Crowd Behavior Analysis}
\label{behavior}
Crowd behavior analysis seeks to understand the dynamics, intent, and patterns of movement within a crowd, enabling proactive management and early detection of hazardous or abnormal situations. It encompasses the recognition of typical group behaviors such as queue formation, lane following, or loitering, as well as the identification of deviations such as panic, sudden dispersal, or aggressive motion. This task is critical for public safety applications, as it provides insight into crowd psychology and supports timely intervention in potentially dangerous scenarios.

LiDAR enables robust behavior analysis through its ability to capture detailed spatiotemporal patterns. By continuously observing changes in position, density, and movement direction, LiDAR data reveals both micro-scale (individual and group) and macro-scale (crowd-level) behaviors. Temporal descriptors, such as displacement vectors and trajectory smoothness, can be used to distinguish between calm and erratic motion. High-density areas with sudden movement variations may indicate crowd turbulence or pushing, while coherent directional flow may reflect organized group behavior.

Crowd anomaly detection can be achieved on LiDAR data by applying regular behavioral assessment baselines and flagging deviations from expected motion or formation patterns~\cite{hattori2023anomaly}. 
These capabilities make LiDAR a tool for understanding crowd behavior, particularly in environments where visibility is limited or individual identification is restricted by privacy constraints.




\vspace{-0.4cm}
\section{AI-LiDAR Crowd Monitoring System}

\begin{figure*}[!ht]
\centering
\includegraphics[width=\textwidth]{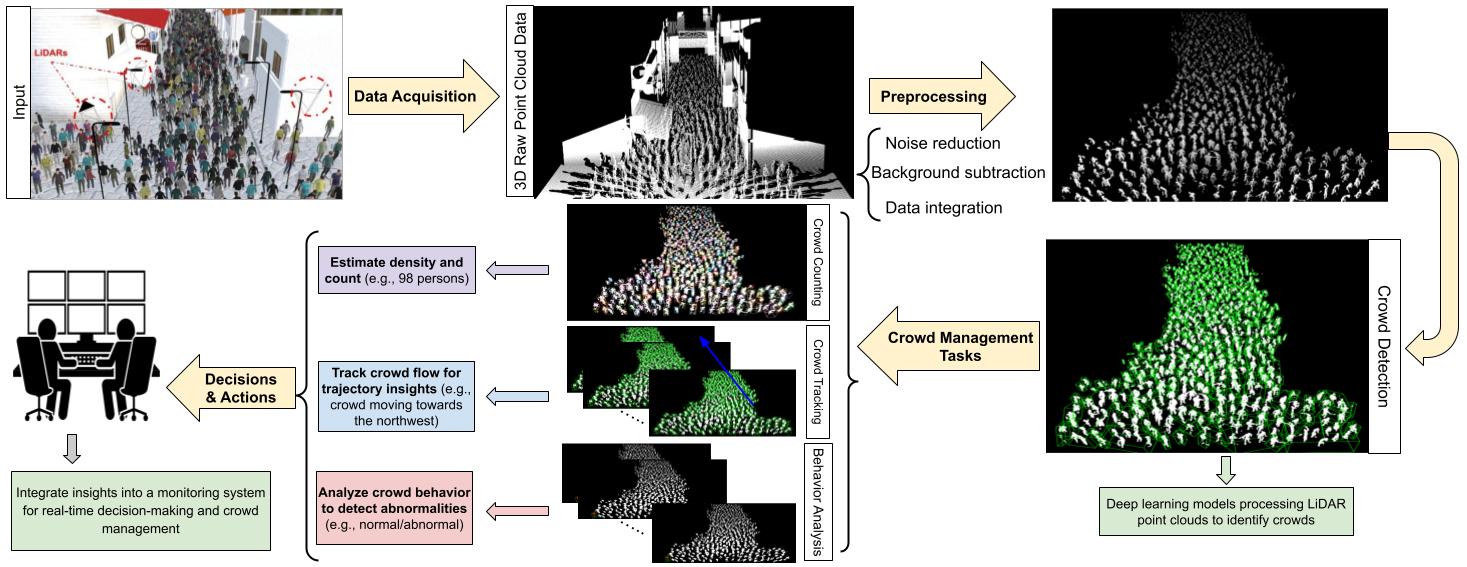}
\caption{\textcolor{black}{End-to-end AI-enabled LiDAR crowd management framework illustrating the main processing stages from data acquisition to crowd analysis tasks and decision support.}}
\label{fig:system_architecture}
\end{figure*}

AI plays a pivotal role in transforming raw LiDAR data into actionable insights for crowd management by analyzing meaningful patterns from high-dimensional, dynamic point clouds. The integration of AI techniques enables automation, improves accuracy, and supports real-time decision-making in complex urban environments.

Fig.~\ref{fig:system_architecture} illustrates a high-level architecture of an AI-enabled LiDAR crowd monitoring system. The system processes begin with LiDAR deployment and data collection, followed by preprocessing and descriptor extraction. These processed point clouds feed into specialized AI models that perform detection, counting, tracking, and behavior classification. The insights generated are then used in a decision support system to take informed actions.

\vspace{-0.4cm}
\subsection{Data Flow and AI-driven Processing}
AI–LiDAR systems deploy sensors on fixed infrastructure or mobile platforms, with optimal placement to achieve maximum area coverage. Raw point cloud data is continuously collected by the deployed sensors. Although it offers high spatial resolution, it requires significant processing before it can be used for high-level crowd management tasks.

The preprocessing stage is a critical component of the pipeline to prepare the data for the main AI components. This step includes noise filtering, background subtraction, and data registration to align point clouds from different frames or sensors, thereby mitigating environmental effects and erroneous measurements. Preprocessing also involves extracting crowd descriptors, including height distributions, surface normals, motion vectors, and reflectivity patterns. 

AI techniques applied to point clouds fall into two categories: handcrafted methods and learning-based approaches. \textcolor{black}{ While handcrafted methods rely on geometric clustering and rule-based heuristics, crowd-specific deep learning architectures are designed to address challenges unique to dense crowd scenarios. These include severe occlusions, overlapping point clusters, and large variations in local density. As a result, recent approaches employ multi-scale feature extraction to handle varying crowd densities, attention mechanisms to improve robustness under occlusion, and specialized loss functions that balance individual-level and collective-level representations. In contrast, general-purpose point cloud models such as PointNet++ may struggle in dense crowds due to hierarchical sampling strategies that merge overlapping individuals, motivating the development of density-aware convolutions and occlusion-robust attention mechanisms tailored to crowd point cloud data.}

The output of this processing stage includes cleaned, structured data enriched with semantic information, ready to be passed into application-specific modules such as crowd tracking, counting, and behavior analysis. By leveraging AI in the processing pipeline, the system gains adaptability to different crowd scenarios and environmental conditions, improving both robustness and scalability.

\vspace{-0.2cm}
\subsection{Informed Decision Making}
The final stage of the AI–LiDAR monitoring system translates crowd analysis outputs into actionable intelligence for operators and automated systems. Insights from different crowd management applications are fed into a centralized decision-making module that supports real-time monitoring, alert generation, and preventive control measures. 3D Visual dashboards display live density maps, motion patterns, and anomaly alerts, enabling human operators to assess the situation at a glance. When abnormal behaviors or hazardous conditions, such as overcrowding, panic, or sudden flow disruptions, are detected, the system can trigger automated alerts to security personnel or activate control mechanisms, such as dynamic signage, barrier deployment, or route redirection.

Moreover, the continuous collection and analysis of crowd data support long-term planning and policy-making. Historical trends can inform infrastructure design, event logistics, and emergency preparedness strategies. By closing the loop between sensing, analysis, and action, AI–LiDAR systems not only enhance situational awareness but also enable data-driven, proactive crowd management in smart cities and large public spaces.


\section{Taxonomy of LiDAR-Based Crowd Management \& Open Research Directions}\label{LiDAR_applications}
In this section, we present a comprehensive taxonomy of LiDAR-based crowd management systems and discuss open research directions to leverage for practical implementation.
\subsection{Taxonomy}

\begin{figure*}[!ht]
\centering
\includegraphics[width=\textwidth]{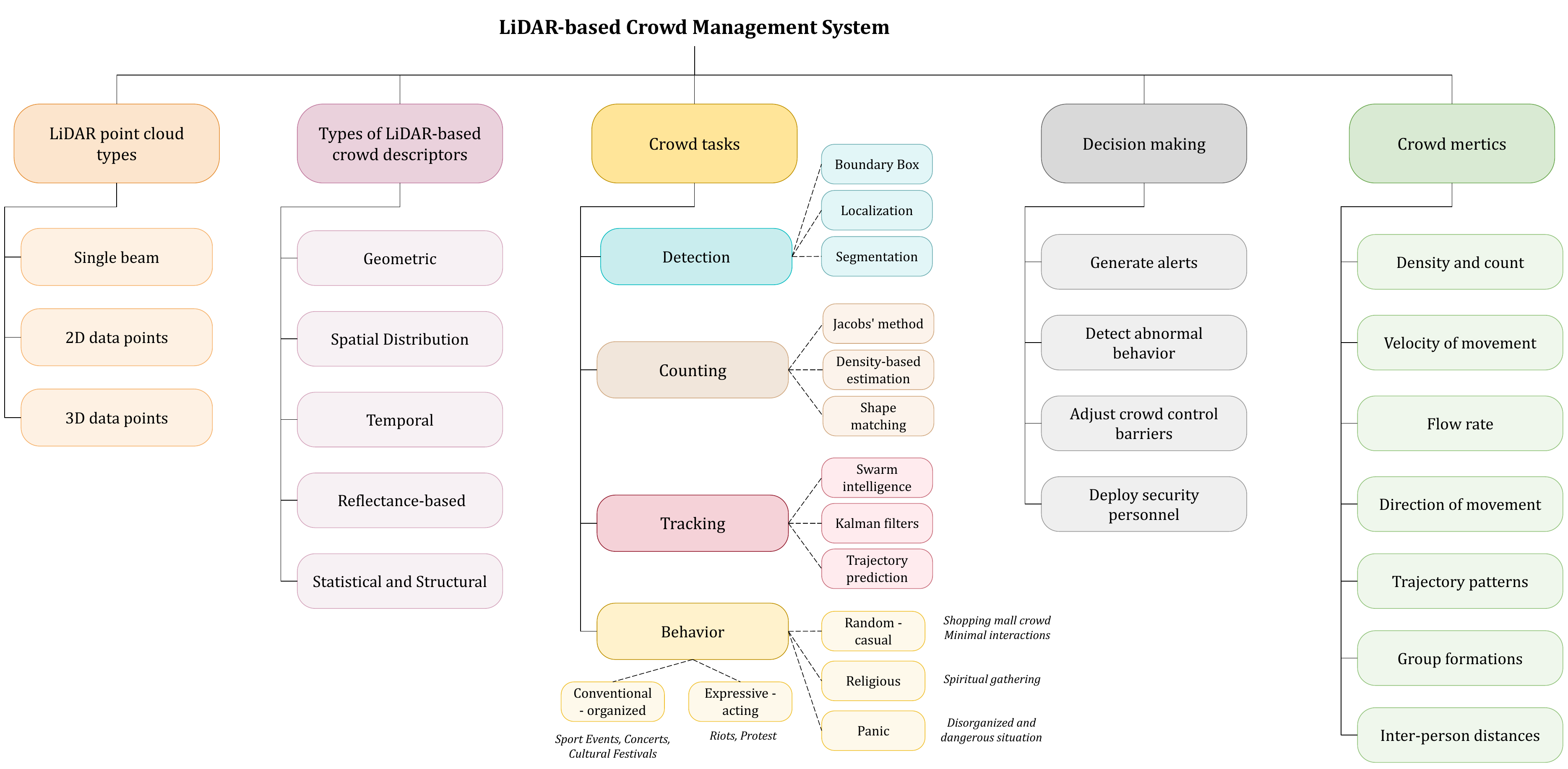}
\caption{\textcolor{black}{High-level, integrated taxonomy of LiDAR-based crowd management systems illustrating the relationships between point cloud types, crowd descriptors, analytical methods, core tasks, decision-making components, and application domains.}}
\label{fig:lidar_taxonomy}\vspace{-0.4cm}
\end{figure*}

The existing literature lacks a comprehensive taxonomy of LiDAR-based crowd management systems. Therefore, we propose a structured framework, depicted in Fig.~\ref{fig:lidar_taxonomy}, based on the inputs presented in this paper to summarize the technical components and application domains, offering a more complete picture of these systems. \textcolor{black}{The elements in Fig.~\ref{fig:lidar_taxonomy} are intentionally presented in a unified view to emphasize their interdependencies within end-to-end LiDAR-based crowd management systems.}

\subsection{Open Research Directions}
\label{sec:directions}





\textbf{Lack of point cloud datasets for crowd analysis:} Despite the growing adoption of LiDAR in various domains, there is a notable shortage of publicly available point cloud datasets specifically designed for crowd management issues. \textcolor{black}{This scarcity significantly impedes the development and benchmarking of AI models for crowd-related tasks. Nevertheless, existing point cloud datasets have been mainly collected for autonomous driving or general scene understanding and only partially capture pedestrian-related scenarios. Representative examples include Waymo~\cite{sun2020scalability}, nuScenes~\cite{caesar2020nuscenes}, ApolloScape~\cite{huang2018apolloscape}, RoboSense~\cite{su2025robosense}, and PandaSet ~\cite{xiao2021pandaset}, which provide annotated LiDAR data for detection or tracking but are limited in terms of crowd density, interaction complexity, and behavioral annotations. Consequently, these datasets can only be repurposed for preliminary evaluation and benchmarking, highlighting a critical gap and an important open research direction: the need for dedicated LiDAR datasets tailored to realistic, high-density crowd scenarios.}

\textcolor{black}{In parallel, synthetic data generation using 3D simulation platforms (e.g., Blender, Carla, or UnrealEngine) enabling LiDAR scanning is increasingly employed to produce controllable crowd scenes with precise ground-truth annotations. While these approaches help alleviate data scarcity, they also highlight the need for continued development of realistic synthetic generation pipelines and standardized benchmarking protocols tailored to LiDAR-based crowd analysis.}

\textbf{LiDAR sensor mounting and placement:} Effective deployment of LiDAR sensors for crowd management requires strategic placement to maximize coverage and data quality while minimizing deployment costs. In both indoor and outdoor environments, achieving complete visibility of crowded areas often depends on sensor height, orientation, and FOV. Fixed infrastructure, such as rooftops, light poles, or ceilings, can provide stable coverage but may require multiple sensors to eliminate blind spots, thereby increasing costs and calibration complexity. Alternatively, mobile platforms, including ground robots or UAVs, provide flexibility to dynamically reposition sensors based on crowd density or event context. UAV-mounted LiDAR systems are particularly valuable for large-scale outdoor events or areas lacking built infrastructure, but they introduce additional challenges such as flight stability, battery life, and regulatory constraints. Determining the optimal placement strategy, i.e., balancing sensor count, mounting height, angle, and platform mobility, is essential for cost-effective, scalable crowd monitoring solutions.

\textcolor{black}{\textbf{Processing, deployment, and integration challenges:} 
Real-time LiDAR-based crowd monitoring poses significant computational challenges due to the high dimensionality of point cloud data, which demands substantial processing resources for filtering, segmentation, and analysis. Edge-based deployments are constrained by memory, energy, and thermal limits, while cloud-based processing introduces latency and bandwidth overheads that can affect real-time responsiveness.}

\textcolor{black}{In parallel, the high cost of high-resolution LiDAR sensors remains a practical barrier to large-scale deployment in public environments such as plazas, transport hubs, or stations. Although LiDAR offers advantages in privacy preservation, robustness to lighting conditions, and accurate 3D perception, these benefits must be carefully weighed against sensor cost, infrastructure requirements, and maintenance effort. Nevertheless, strategic sensor placement and coverage-aware deployment can partially mitigate costs, as fewer LiDAR units may be required compared to alternative sensing modalities due to their extended range and wide field of view. Additional economic justification can be achieved through reduced personnel needs and the prevention of safety-critical incidents. Hybrid edge–cloud architectures, combined with data compression and adaptive workload distribution, are therefore key to achieving scalable, cost-effective LiDAR-based crowd-monitoring systems.}

\textcolor{black}{Finally, integrating LiDAR with complementary sensing modalities such as RGB cameras, thermal sensors, and radar can further improve monitoring accuracy and robustness. While LiDAR provides precise geometric information, other sensors contribute contextual cues such as appearance and temperature. Effective multimodal fusion requires accurate calibration and cross-modal feature representation for real-world applications.} 

\textbf{Generative AI for 3D crowd management:} While large multimodal and vision–language models are rapidly evolving, their application to 3D point cloud understanding remains relatively underexplored. \textcolor{black}{Generative models can help mitigate the data scarcity challenge at two complementary levels. First, representation-level generative models, such as diffusion-based architectures, can learn robust latent embeddings of LiDAR point clouds in self-supervised or weakly supervised settings, reducing reliance on large annotated datasets. Second, generative models can be integrated into simulation-to-real pipelines to synthesize realistic LiDAR crowd scenes with controllable density, occlusion patterns, and motion behaviors, thereby augmenting limited real-world data and enabling safer evaluation of rare or hazardous scenarios. Beyond data generation, multimodal prompting that combines text, images, and 3D point clouds could support intuitive, human-centered interaction with crowd monitoring systems. Realizing this vision requires addressing challenges related to efficient 3D embeddings, temporal coherence, real-time deployment constraints, and the development of domain-specific foundation models for privacy-aware 3D crowd analytics.}

\section{Conclusion}
This paper explored the use of LiDAR technology for crowd management, focusing on its advantages over conventional sensors, key analytical tasks such as detection, counting, tracking, and behavior analysis, and the integration of AI for real-time decision-making. LiDAR’s ability to capture high-resolution, privacy-preserving 3D spatial data makes it a powerful tool for understanding and managing crowds in complex environments. However, despite its strong potential, LiDAR-based crowd monitoring remains an emerging field. Key challenges, including data availability, processing complexity, and system integration, must be addressed before the technology reaches full maturity. Continued research and development are essential to unlock its capabilities in real-world applications.

\section{Acknowledgment}
\label{sec:acknowledgment}

This work was supported, in part, by the Center to Stream Healthcare in Place (C2SHiP), an NSF Industry-University Cooperative Research Center (IUCRC) under Grant \#2052528.

\bibliographystyle{IEEEtran}
\bibliography{references}

\balance
\newpage
 
\vspace{11pt}

\vfill

\end{document}